\documentclass[runningheads]{llncs}
\RequirePackage{amsmath,amssymb}
\usepackage{graphicx}
\usepackage{hyperref}
\usepackage{color}
\usepackage[width=122mm,left=12mm,paperwidth=146mm,height=193mm,top=12mm,paperheight=217mm]{geometry}

\usepackage{xspace}

\newcommand{\figref}[1]{\mbox{Figure~\ref{#1}}}
\newcommand{\tabref}[1]{\mbox{Table~\ref{#1}}}

\makeatletter
\DeclareRobustCommand\onedot{\futurelet\@let@token\@onedot}
\def\@onedot{\ifx\@let@token.\else.\null\fi\xspace}

\def\ie{\emph{i.e}\onedot} 
 
\def\etc{\emph{etc}\onedot} 
 
\def\etal{\emph{et al}\onedot}
\makeatother

\begin{document}
\pagestyle{headings}
\mainmatter

\title{Learning to See the Invisible: End-to-End Trainable Amodal Instance Segmentation} 

\titlerunning{Learning to See the Invisible}

\authorrunning{P. Follmann \etal}

\author{Patrick Follmann, Rebecca K\"onig, Philipp H\"artinger, Michael Klostermann}


\institute{MVTec Software GmbH, \\
  www.mvtec.com, \\
	\email{ \{follmann,koenig,haertinger,klostermann\}@mvtec.com}
}

\maketitle

\begin{abstract}
Semantic amodal segmentation is a recently proposed extension to instance-aware segmentation that includes the prediction of the invisible region of each object instance.
We present the first all-in-one end-to-end trainable model for semantic amodal segmentation that predicts the amodal instance masks as well as their visible and invisible part in a single forward pass.

In a detailed analysis, we provide experiments to show which architecture choices are beneficial for an all-in-one amodal segmentation model.
On the \emph{COCO amodal} dataset, our model outperforms the current baseline for amodal segmentation by a large margin. To further evaluate our model, we provide two new datasets with ground truth for semantic amodal segmentation, \emph{D2S amodal} and \emph{COCOA cls}. For both datasets, our model provides a strong baseline performance.
Using special data augmentation techniques, we show that amodal segmentation on \emph{D2S amodal} is possible with reasonable performance, even without providing amodal training data.

\end{abstract}

\begin{figure*}
  \centering
  \includegraphics[width=0.3\textwidth]{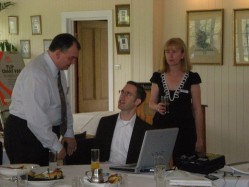}
  \includegraphics[width=0.3\textwidth]{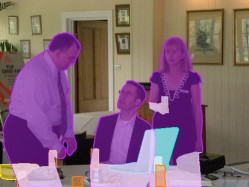}
  \includegraphics[width=0.3\textwidth]{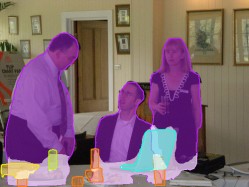}
  \caption{\textbf{Learning the Invisible.}
    (\emph{Left}) input image;
    (\emph{middle}) ground-truth amodal instance annotations;
    (\emph{right}) predictions of our model: amodal instance masks and occlusion masks.
    The mask color encodes the object class and occlusion masks are highlighted in light color
  }
  \label{fig:eyecatcher}
\end{figure*}

\section{Introduction}

Humans, with their strong visual system, have no difficulties reasoning about foreground and background objects in a two-dimensional image. At the same time, humans have the ability of \textit{amodal perception}, \ie to reason about the invisible, occluded parts of objects \cite{kellman2013perceptual,lehar1999gestalt}. 
Robots that should navigate in their environment and pick or place objects need to know if the objects are occluded or hidden by one or several other instances.
This problem leads to the task of semantic amodal segmentation, \ie, the combination of segmenting each instance within an image by predicting its amodal mask and determining which parts of the segmented instances are occluded and what the corresponding occluder is.

The amodal mask is defined as the union of the visible mask (which we will also refer to as modal mask as opposed to the amodal mask) and the invisible occlusion mask of the object. Predicting amodal and visible masks simultaneously provides a deeper understanding of the scene. For example, it allows to calculate regions of occlusion and lets the robot know which objects have to be removed or in which direction to move in order to get free access to the object of interest.

As shown in the example of \figref{fig:eyecatcher}, we not only want to reason about the visible parts of objects in the image, but also want to predict the extension of the objects into their occluded, invisible parts. We propose a model that can predict the visible, invisible, and amodal masks for each instance simultaneously without much additional computational effort.

Predicting the invisible part of an object is difficult: If the object is occluded by an object from another category, the model has no visual cues how to extend the visible mask into the occluded object part. There are generally no edges or other visual features that indicate the contour of the occluded object. In contrast, if the object is occluded by another instance of the same category, it is very hard for the model to judge where to stop expanding the mask into the occlusion part as the category-specific features are present all around.

Note that throughout the paper, we will call annotations containing only visible masks and models predicting visible masks \emph{modal}, in contrast to \emph{amodal} annotations and methods. We will also use the terms occlusion masks and invisible masks as synonyms.

In summary, our paper contains the following contributions:
\begin{itemize}
  \item To the best of our knowledge, we are the first to propose an all-in-one, end-to-end trainable multi-task model for semantic segmentation that simultaneously predicts amodal masks, visible masks, and occlusion masks for each object instance in an image in a single forward pass.
  \item We provide the new semantic amodal segmentation dataset \emph{D2S amodal}, which is based on \emph{D2S} \cite{d2s2018}, with guaranteed annotation-completeness and high-quality annotations including amodal, visible, and invisible masks. Additionally, it contains information about the depth ordering of object instances. In comparison to the class-agnostic \emph{COCO amodal} dataset \cite{zhu2017semantic}, \emph{D2S amodal} contains 60 different object categories and allows to predict amodal and occlusion masks class-specifically.
  \item By merging the categories of the modal \emph{COCO} dataset to the instances of \emph{COCO amodal} we obtain the new amodal dataset \emph{COCO amodal cls} with class labels.
  \item We show that our architecture outperforms the current baseline on \emph{COCO amodal} \cite{zhu2017semantic} and set a strong baseline on \emph{D2S amodal}. We provide extensive evaluations in order to compare different architectural choices for training.
  \item The training set of \emph{D2S} allows to apply extensive data augmentation. This makes it possible to train a semantic amodal method without any amodally annotated data.
\end{itemize}

\section{Related Work}

The topic of amodal perception has already been addressed in various fields of computer vision research.

\paragraph{Semantic Segmentation and 3D Scene Reconstruction.} Two tasks, where amodal completion has already been used for some years, are semantic segmentation and 3D scene reconstruction. The task of semantic segmentation is to predict a category label for each pixel in an image. Semantic segmentation does not take different object instances into account, but returns a single region for each of the possible classes.
Classes are often related to background or \emph{stuff}, such as \emph{sky, water, ground, wall,} \etc.
In \cite{guo2012beyond}, Guo and Hoiem describe a method to infer the entire region of occluded background surfaces. Their algorithm detects the occluding objects and fills their regions with the underlying or surrounding surface.

In 3D reconstruction, parts of the scene can often not be reconstructed because of occlusions.
Gupta \etal \cite{gupta2013perceptual} combine depth information, superpixels, and hierarchical segmentations for amodal completion of semantic surfaces. Also Silberman \etal \cite{silberman2014contour} address the problem of surface completion in the setting of a 2.5D sensor. They use a conditional random field in order to complete contours. The completed contours are subsequently used for surface completion.

In contrast to the above mentioned semantic segmentation methods, our work does not deal with the amodal completion of background regions or 3D object surfaces, but focuses on object instances in 2D images.

\paragraph{Object Detection.} In the context of object detection, Kar \etal \cite{kar2015amodal} use a CNN to predict amodal bounding boxes of objects.
By additionally estimating the depth of the bounding boxes and the focal length of the camera, object dimensions can be derived. 
However, neither the object mask nor the occluded part of the object is predicted.

\paragraph{Instance Segmentation.} More recent methods extend the \emph{object detection} task to the more challenging \emph{instance segmentation} task to predict the category and visible segmentation mask of each object instance in an image \cite{li_2016_fully,he_2017_mask,li2016iterative}.
Yang \etal \cite{yang2012layered} propose a probabilistic model that uses the output of object detectors to predict instance shapes and their depth ordering. However, no occlusion regions are predicted. In \cite{chen2015multi}, Chen \etal propose a graph-cut algorithm with occlusion handling in order to improve the quality of visible masks. However, they do neither predict occlusion nor amodal masks.

\paragraph{Amodal Instance Segmentation.} Research on amodal instance segmentation or semantic amodal segmentation has just started to emerge. 
Li and Malik \cite{li2016amodal} were the first to provide a method for amodal instance segmentation. They extend their instance segmentation approach \cite{li2016iterative} by iteratively enlarging the modal bounding box of an object into the directions of high heatmap values and recomputing the heatmap.
Due to the lack of amodal instance segmentation ground truth, they use modally annotated data and data augmentation in order to train and evaluate their model.
In \cite{zhu2017semantic}, Zhu \etal provide a new and pioneering dataset \emph{COCO amodal} for amodal instance segmentation based on images from the original \emph{COCO} \cite{lin_2014_coco} dataset. The authors did not restrict the annotations to the usual \emph{COCO} classes and annotators could assign arbitrary names to the objects. Therefore, all objects in the dataset belong to a single class \emph{object} and the variety of objects in this class is very large. Additionally, the authors provide annotations of background regions, which are sometimes extending to the full image domain, labeled as \emph{stuff}.
In order to provide a baseline, Zhu \etal use AmodalMask, which is the SharpMask~\cite{pinhero2015} model trained on the amodal ground truth. The model suggests object candidates with a relatively high recall. However, the predictions of the model are class-agnostic. They also trained a ResNet-50 \cite{he2016deep} to predict the foreground object given two input object masks and the corresponding image-patches.

In contrast to \cite{li2016amodal} and \cite{zhu2017semantic}, our model is class-specific, end-to-end trainable, lightweight, and can predict amodal, visible, and invisible instance masks in a single forward-pass.

\section{End-to-End Architecture for Prediction of Amodal, Visible, and Invisible Masks}

\subsection{Architecture}

\begin{figure}[!t]
  \centering
  \includegraphics[trim = 1mm 1mm 1mm 25mm, clip, width=0.9\textwidth]{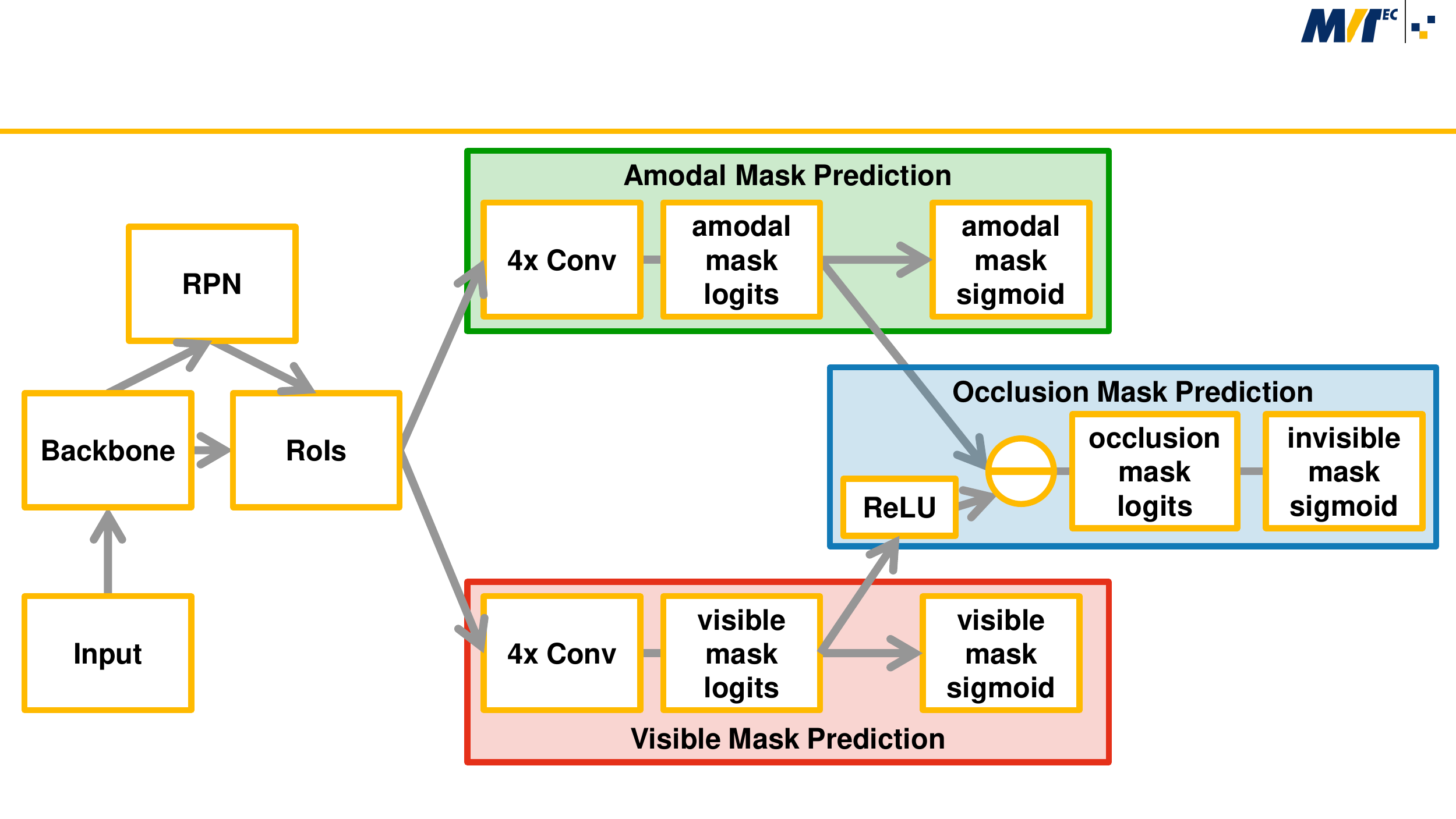} 
  \caption{\textbf{ORCNN architecture.}}
  \label{fig:architecture}
\end{figure}

We name our method \emph{Occlusion R-CNN} (\emph{ORCNN}), as its architecture is based on Mask R-CNN \cite{he_2017_mask} (MRCNN). In the ORCNN architecture we extend MRCNN with additional heads for the prediction of amodal masks (amodal mask head) and the occlusion mask (occlusion mask head). An overview of the ORCNN architecture is shown in \figref{fig:architecture}.

The visible mask head and the amodal mask head share the same architecture and use four $3\times3$ convolutions and ReLU layers to generate meaningful features for mask prediction. Their inputs are the extracted features from the RoIAlign~\cite{he_2017_mask} layer. Note that during training and inference, the amodal and visible mask prediction heads of ORCNN share the same box proposals generated by the region proposal network (RPN). The target ground truth masks of the RPN are the bounding boxes of the amodal instances. Therefore, the visible mask prediction head has to predict the visible mask of an instance from the amodal bounding box. This is a major difference to a modal model that is trained using the bounding boxes of the modal, \ie visible masks of the instances.

A key component is that we link the modal and amodal mask heads with the occlusion mask head.
The occlusion mask head essentially subtracts the visible from the amodal mask logits in order to obtain the occlusion mask logits. 
It is crucial to apply a ReLU-operation on the visible mask logits before subtraction to avoid occlusion mask prediction for pixels where neither the amodal nor the modal mask are predicted. 

Mounting both the modal, as well as the amodal head on the same RoI-feature extraction module leads to several advantages: First, this makes the additional amodal and occlusion mask prediction light-weight as only five additional convolution modules and two sigmoid layers are necessary. Second, using the same RoIs for the amodal and visible mask prediction guarantees that both predicted masks correspond to the same object prediction. In comparison, if one used two separate models for amodal and visible mask predictions, it would not be straight-forward to fuse the results of these models. And third, by sharing the same architecture for the amodal and visible mask prediction head, we can initialize both heads with the same weights that have been pre-trained on a large modal instance-segmentation dataset, such as \emph{COCO}.

\subsection{Training}

In order to obtain meaningful predictions for the visible, invisible, and occlusion masks, we have to formulate the corresponding losses for each of the tasks. As the tasks are similar and only the ground truth differs, we use a similar sigmoid-cross-entropy loss for all three mask types: First applying a per-pixel sigmoid and thereafter an average binary cross-entropy loss like in \cite{he_2017_mask}.
In combination with the losses for the class (\emph{cls}) and bounding box (\emph{box}), we obtain the total loss $L$:
\begin{equation}
L = L_{\mathit{cls}} + L_{\mathit{box}} + L_{\mathit{AM}} + L_{\mathit{VM}} + L_{\mathit{IVM}},
\end{equation}
where \emph{AM, VM, IVM} are abbreviations for amodal, visible, and invisible mask, respectively. In theory, one of the three losses $L_{\mathit{AM}}, L_{\mathit{VM}}$ and $L_{\mathit{IVM}}$ is redundant, as for ground truth we have $\mathit{IVM} = \mathit{AM} - \mathit{VM}$. Nevertheless, adding an additional loss for occlusion masks leads to amodal mask logits and visible mask logits that are on the same scale. Otherwise, consider the case that for a pixel we have high probability for the amodal and the visible mask. E.g., let the logit activations be 14 and 10, respectively. This leads to a probability of $1/(1 + e^{-4}) = 0.982$ for the occlusion mask to be present at this pixel, although occlusion should not be predicted.
In the same manner, leaving out the loss for the visible mask $L_{\mathit{VM}}$, leads to reasonable results for the occlusion mask but bad results for the visible mask. On the other hand, one could argue that instead of subtracting logits, it would make more sense to multiply or subtract probabilities directly. We have tried that in our experiments as well, but observed that adding a loss acting on the probabilites after computing the sigmoids is numerically unstable.

\subsection{Evaluation}

To judge which model is the best for the task of amodal instance segmentation, we propose extending the mean average precision ($AP$) and mean average recall ($AR$) evaluation measures commonly used for  instance segmentation, e.g. on \emph{Pascal VOC} \cite{everingham_2015_pascal} and \emph{COCO} \cite{lin_2014_coco} benchmarks. For brevity, in the following we will describe the extension of the measure only for the case of $AP$. The extension for $AR$ is straight-forward. As in the \emph{COCO} benchmark, we compute the final $AP$ by averaging the per-category $AP$s, which are in turn the average over ten equally spaced intersection over union (IoU) thresholds from 0.5 to 0.95.

In order to evaluate the individual tasks, we calculate the AP values independently to obtain $AP_{\mathit{A}}$ and  $AP_{\mathit{V}}$ using amodal or visible masks, respectively. We can then include both of the masks into the definition of a true positive instance to obtain a combined $AP$ measure $AP_{\mathit{AV}}$ (amodal-visible AP). For example, in order to obtain a true positive result in the $AP_{\mathit{AV}}$ setting for a given IoU threshold $t$, we need the correct predicted class and additionally, $\mathrm{IoU}(AM^G, AM^P) > t$ and $\mathrm{IoU}(\mathit{VM}^G, \mathit{VM}^P) > t$ both need to be satisfied. Here, $\mathrm{IoU}(AM^G, AM^P)$ is the intersection over union between the amodal mask ground truth $AM^G$ and the amodal mask prediction $AM^P$.

The invisible masks are included only indirectly into the overall measure $AP_{\mathit{AV}}$ due to the following issues:
First, for non-occluded objects, the invisible mask is not present and it is not straight-forward to define recall on something that is not present\footnote{Generally, recall is increasing with more proposals. In case of an empty invisible mask this is not the case, which prevents the usual way of computing the $AP$ measure.}. Second, for most objects in \emph{COCO amodal} or \emph{D2S amodal}, the invisible mask areas are rather small compared to the amodal or visible masks. Hence, small differences in invisible mask predictions have a large influence on the $\mathrm{IoU}$. 
This leads to misleading results of a measure taking into account all three mask types, such as the amodal-invisible-visible average precision $AP_{\mathit{AIVV}}$. For $AP_{\mathit{AIVV}}$, a proposal is only a true positive if all three masks (amodal, visible and invisible) exceed the IoU-threshold. 

To measure the quality of predicted occlusions, we do separate evaluations, where we ignore all non-occluded ground truth objects. For these evaluations, we also calculate the average precision of invisible masks $AP^{0.5}_\mathit{IV}$ at IoU-threshold 0.5. We use a low IoU threshold as the invisible masks are often of very small size and to take the difficulty of the task into account.
For the computation of $AP_\mathit{V}$ and $AP_\mathit{IV}$ we ignore the proposals whose amodal masks have an $\mathrm{IoU}$ with an ignored ground truth of at least 0.5. By this heuristic, we consider only proposals that are meant for occluded ground truth objects.

For models that do not predict any visible or invisible masks, we calculate the measures $AP_{\mathit{AV}}$ and $AP_{\mathit{V}}$ by using amodal masks also as predictions for the visible masks.

\section{Experiments}

In the following, we compare our models to previous results on \emph{COCO amodal} and set new benchmarks for the new semantic amodal datasets \emph{COCO amodal cls} and \emph{D2S amodal}. All models were trained using the Detectron \cite{Detectron2018} framework. More information on the settings is in the supplementary material.

\subsection{\emph{COCOA}}

\emph{COCO amodal} (\emph{COCOA}) \cite{zhu2017semantic} is the first dataset with ground truth for semantic amodal segmentation. The dataset consists of 2500 training, 1323 validation, and 1250 test images. In each image, most objects and background \textit{stuff} areas are annotated with amodal masks. Occluded objects are additionally annotated with visible and invisible masks. Furthermore, a binary depth-ordering for the annotated objects is provided. All objects belong to a single category \textit{object} and have an additional \textit{stuff} label.

\paragraph{Amodal Mask Prediction.} As a baseline result for amodal semantic segmentation we train MRCNN with a ResNet-50 or ResNet-101 backbone on the amodal annotations. We call these models \emph{AmodalMRCNN-50/-101}. \tabref{tbl:coco_amodal_baseline} compares AmodalMRCNN to the baseline of \cite{zhu2017semantic} using their evaluation tool. AmodalMRCNN outperforms AmodalMask by a large margin in terms of average precision. AmodalMask achieves a high recall, since it always predicts 1000 regions for each image. Nevertheless, AmodalMRCNN achieves even higher recall while predicting only 30 results per image on average. An exception is the category \emph{stuff}, since \emph{stuff} regions are often very large areas in the background of the image and AmodalMRCNN predicts no object proposals for these areas.

\begin{table}
  \centering
  \caption{{\bf Baseline results for \emph{COCOA}.}
    Amodal mean average precision and average recall values for AmodalMask 
    \cite{zhu2017semantic} compared to AmodalMRCNN}
  \label{tbl:coco_amodal_baseline}
  \footnotesize
  \def\arraystretch{1.2}
  \setlength{\tabcolsep}{8pt}
  \begin{tabular}{l|cc|cc|cc}
    \hline
    & \multicolumn{2}{c|} {\textbf{all}} & \multicolumn{2}{c|} {\textbf{things}} & \multicolumn{2}{c}{\textbf{stuff}} \\
    & $AP_{\mathit{A}}$ & $AR_{\mathit{A}}$ & $AP_{\mathit{A}}$ & $AR_{\mathit{A}}$ & $AP_{\mathit{A}}$ & $AR_{\mathit{A}}$ \\
    \hline\hline
    AmodalMask \cite{zhu2017semantic} &  5.7  & 43.4  &  5.9  &  45.8 &  0.8 & \textbf{36.7} \\
    AmodalMRCNN-50 & \textbf{29.9}  & \textbf{45.8}  & \textbf{33.2}  &  \textbf{50.4} &  \textbf{5.8} & 33.0 \\
    \hline
  \end{tabular}
\end{table}

In the following, our focus is on \emph{things}, as masks for \emph{stuff} are hard to define and therefore, the variance of annotations between different annotators is high. Thus, in \emph{COCOA no stuff}, we exclude \emph{stuff} annotations during training and evaluation. We found that for AmodalMRCNN and ORCNN $AR$ is generally in line with the $AP$ since recall is already captured within the $AP$ measure. Therefore, for the following evaluations we will just show $AP$ values. In order to highlight the performance on occluded objects, we also evaluate the architectures ignoring all non-occluded instances.

\paragraph{Data augmentation.} We found that on \emph{COCOA}, AmodalMRCNN overfits to the training data already after one epoch. A common way to handle overfitting is to enlarge the training set. Similar to \cite{li2016amodal} and \cite{zhu2017semantic}, we generate new artificial training data by randomly sampling images from the original \emph{COCOA} training set and overlaying one \textit{no-stuff}-object with a \textit{no-stuff}-object from another image. However, this synthetically augmented training data does not improve the performance. This outcome is in line with the results of \emph{AmodalMask}$^\mathit{S}$ from \cite{zhu2017semantic}. It seems to be important for the model to have all objects in context. Furthermore, we generate a new amodal training set by pasting objects from the original \emph{COCO} dataset into randomly sampled \emph{COCO} images. Thus, in this dataset only the available modal annotations from \emph{COCO} are used. Amodal annotations, including occlusion masks, are only present for those objects that are occluded by one of the pasted objects.
In the evaluation, object categories are ignored since they are not provided in \emph{COCOA val}. 
We found that also on \emph{COCO} the described data augmentation does not help. The results for data augmentation are summarized in \tabref{tbl:COCOA_aug_results}.
\vspace{-.2cm}
\paragraph{Occlusion Prediction.} To set a baseline for a model like ORCNN that can predict amodal, visible, and invisible masks at the same time, we use AmodalMRCNN. As the model does not explicitly predict visible masks, we use the amodal masks also as visible mask predictions in the evaluation. 
The results on \emph{COCOA} are summarized in \tabref{tbl:COCOA_results}. Note that in case of non-occluded objects, the amodal mask is the same as the visible mask and therefore models without explicit visible and invisible mask predictions profit on these instances. The multi-task model ORCNN improves the quality of visible masks compared to AmodalMRCNN and sometimes even compared to MRCNN, while at the same time predicting occlusion masks. The results for amodal masks are slightly worse than compared to the single task model AmodalMRCNN.

\begin{table}[ht]
  \caption{\textbf{Data augmentation on \emph{COCOA}.} Models with $^*$ are trained with artificially generated images based on \emph{COCOA}. Models with $^{**}$ are trained with artificially generated images based on \emph{COCO}.}
  \label{tbl:COCOA_aug_results}
  \centering
  \small
  \bgroup
  \def\arraystretch{1.2} 
  \setlength{\tabcolsep}{4pt}
  \begin{tabular}{l | r r r | r r r} \hline
    & \multicolumn{3}{|c|}{all} & \multicolumn{3}{c}{occluded} \\
    dataset \& model & $AP_{\mathit{AV}}$ & $AP_{\mathit{A}}$ & $AP_{\mathit{V}}$ & $AP_{\mathit{AV}}$ & $AP_{\mathit{A}}$ & $AP_{\mathit{V}}$ \\ \hline \hline
    AmodalMRCNN-101        & \textbf{27.8} & \textbf{35.6} & \textbf{29.4} & \textbf{14.7} & \textbf{25.4} & 18.5 \\
    AmodalMRCNN-101$^*$    & 27.2 & 34.7 & 28.9 & 13.3 & 24.6 & 17.5 \\
    AmodalMRCNN-101$^{**}$ & 21.9 & 24.1 & 26.8 & 11.2 & 14.5 & \textbf{19.1} \\
    \multicolumn{7}{c}{} \hfill
  \end{tabular}
  \egroup
\end{table}


\begin{table}[ht]
  \caption{\textbf{\emph{COCOA} results.} Note that only ORCNN is predicting visible and invisible masks additionally to the amodal masks. For all other models, the predicted amodal mask was used for the evaluation of $\mathit{AV}$ and $\mathit{V}$ measures}
  \label{tbl:COCOA_results}
  \centering
  \small
  \bgroup
  \def\arraystretch{1.2} 
  \setlength{\tabcolsep}{4pt}
  \begin{tabular}{l | r r r | r r r r} \hline
    & \multicolumn{3}{|c|}{all} & \multicolumn{4}{c}{occluded} \\
    dataset \& model & $AP_{\mathit{AV}}$ & $AP_{\mathit{A}}$ & $AP_{\mathit{V}}$ & $AP_{\mathit{AV}}$ & $AP_{\mathit{A}}$ & $AP_{\mathit{V}}$ & $AP^{0.5}_{\mathit{IV}}$ \\ \hline \hline
    \emph{COCOA} &  & & & & & \\
    \hspace*{5mm}AmodalMask \cite{zhu2017semantic} & 3.7 & 5.7 & 4.8 & 1.2 & 3.2 & 2.6 & - \\
    \hspace*{5mm}MRCNN-101              & 17.0 & 18.6 & 21.7 & 8.6 & 10.6 & 15.3 & - \\
    \hspace*{5mm}AmodalMRCNN-101        & \textbf{24.1} & \textbf{31.3} & 26.1 & \textbf{11.8} & \textbf{21.6} & 16.2 & - \\
    \hspace*{5mm}ORCNN                  & 21.2 & 25.7 & \textbf{26.7} & 11.5 & 17.0 & \textbf{18.4} & \textbf{1.3} \\ \hline 
    \emph{COCOA no stuff}  & & & & & & \\
    \hspace*{5mm}MRCNN-101              & 22.0 & 23.9 & 27.9 & 12.4 & 15.0 & \textbf{21.6} & - \\
    \hspace*{5mm}AmodalMRCNN-101        & \textbf{27.8} & \textbf{35.6} & 29.4 & \textbf{14.7} & \textbf{25.4} & 18.5 & - \\
    \hspace*{5mm}ORCNN                  & 25.1 & 30.1 & \textbf{30.0} & 14.3 & 20.8 & 21.4 & \textbf{3.0} \\ 
    \multicolumn{8}{c}{} \hfill
  \end{tabular}
  \egroup
\end{table}

\subsection{\emph{COCOA cls}}

For amodal completion the model has to get some intuition about the common shape of objects. We evaluate whether the prediction of amodal masks is a class-specific task. Therefore, we generated a new dataset \emph{COCOA cls} by merging the usual \emph{COCO} 2014 annotations with the \emph{COCOA} dataset. On the one hand, \emph{COCOA} contains many objects of categories (e.g. sandals, sneakers, or stuff categories) that are not part of \emph{COCO}. Although, each object in \emph{COCOA} has a name tag, the annotators were free to choose this name. Furthermore, not all objects present in the ground truth of \emph{COCO} have been annotated in \emph{COCOA}. Therefore, to assign a class-label to the objects in \emph{COCOA}, we calculate the $\mathrm{IoU}$s of the visible masks with the masks given for the corresponding image-id in \emph{COCO}. Only annotations, for which the $\mathrm{IoU}$ between visible mask and any \emph{COCO} annotation exceeds a threshold of 0.75 (and not labeled as stuff or crowd), were kept for \emph{COCOA cls}. Overall, \emph{COCOA cls} has 3,501 images with 10,592 objects compared to the 3,823 images and 34,916 objects of \emph{COCOA}. Note that using this merging scheme, some \emph{COCO} classes, e.g. \emph{hairdryer}, are not present in the training set of \emph{COCOA cls}. Furthermore, for many images not all \emph{COCO} annotations can be matched to a corresponding \emph{COCOA} label.


\begin{figure*}
  \centering
  \includegraphics[width=0.19\textwidth]{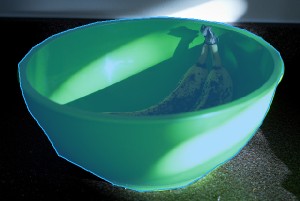}
  \includegraphics[width=0.19\textwidth]{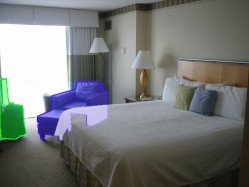}
  \includegraphics[width=0.19\textwidth]{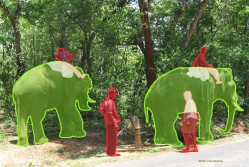}
  \includegraphics[width=0.19\textwidth]{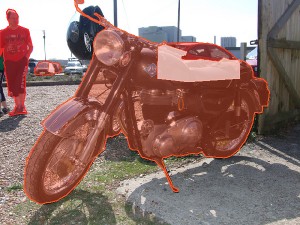}
  \includegraphics[width=0.19\textwidth]{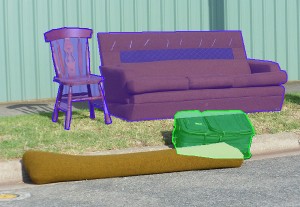}\\
  \includegraphics[width=0.19\textwidth]{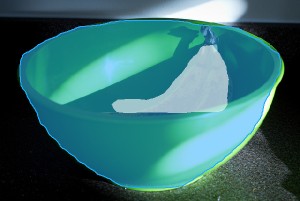}
  \includegraphics[width=0.19\textwidth]{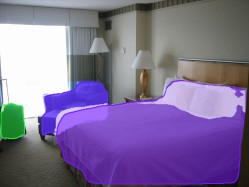}
  \includegraphics[width=0.19\textwidth]{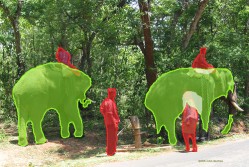}
  \includegraphics[width=0.19\textwidth]{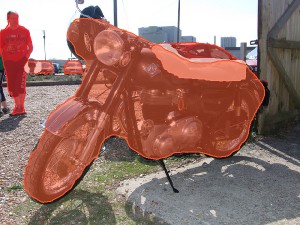}
  \includegraphics[width=0.19\textwidth]{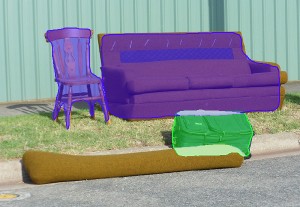}
  \caption{\textbf{\emph{COCOA cls} results.} \textit{top:} ground truth annotations for images of the validation set; \textit{bottom:} exemplary qualitative results of our ORCNN model trained on the \texttt{train} split. Predicted invisible masks are indicated by a white overlap. Note that ORCNN sometimes predicts instances or invisible masks correctly that are not part of the ground truth. These count as false positives in the evaluation or at least lead to reduced $AP_\mathrm{AV}$ and $AP_\mathrm{V}$ values.}
  \label{fig:coco_cls_results}
\end{figure*}

As shown in \tabref{tbl:COCOA_cls_results}, models perform a lot better on \emph{COCOA cls} compared to \emph{COCOA}. AmodalMRCNN, a model that does not predict any visible or invisible masks, outperforms ORCNN on \emph{COCOA cls}. There are several reasons to explain this behavior: First, there could be a bias towards objects with no or only minor occlusions. Second, the quality of visible masks predicted by ORCNN is slightly worse than just using the amodal masks predicted by AmodalMRCNN. ORCNN tries to predict invisible masks as soon as objects are touching each other. As there are three instead of only one output mask compared to AmodalMRCNN, ORCNN has more sources of error. Therefore, a direct comparison is difficult. For visible masks, the best results are obtained with MRCNN. This makes sense, as the model was trained to predict visible masks on \emph{COCO}.

In order to see if class-specific mask prediction improves the results, AmodalMRCNN and ORCNN were also trained using class-agnostic mask proposals. On \emph{COCOA cls}, using class-specific mask proposals helps for almost all measures both in the case of AmodalMRCNN and ORCNN. For occluded objects, class-agnostic ORCNN gets the best average precisions in terms of visible and invisible masks.

Generally, the interpretation of $AP$ values for \emph{COCOA} and \emph{COCOA cls} results is difficult because for both datasets the annotations are not complete. For example, as shown in \figref{fig:coco_cls_results}, there are some cases where ORCNN makes correct predictions but the ground truth does not contain the corresponding annotations.

\begin{table}[ht]
  \caption{\textbf{\emph{COCOA cls} results.} Models marked with (cls-agn) are using class-agnostic mask prediction heads.}
  \label{tbl:COCOA_cls_results}
  \centering
  \small
  \bgroup
  \def\arraystretch{1.2} 
  \setlength{\tabcolsep}{4pt}
  \begin{tabular}{l | r r r | r r r r} \hline
    & \multicolumn{3}{|c|}{all} & \multicolumn{4}{c}{occluded} \\
    dataset \& model & $AP_{\mathit{AV}}$ & $AP_{\mathit{A}}$ & $AP_{\mathit{V}}$ & $AP_{\mathit{AV}}$ & $AP_{\mathit{A}}$ & $AP_{\mathit{V}}$ & $AP^{0.5}_{\mathit{IV}}$ \\ \hline \hline
    MRCNN-101 \cite{he_2017_mask} & 39.0 			& 39.8 			& \textbf{44.9}		& \textbf{25.4} & 26.5 & \textbf{34.9} & - \\
    AmodalMRCNN-101 (cls-agn) 	& 37.1 			& 40.4 			& 38.6 			& 23.9 & 28.7 & 26.8 & - \\
    AmodalMRCNN-101 			& \textbf{38.8} & \textbf{41.7} & 40.5 & 24.9 & \textbf{29.2} & 28.0 & - \\
    ORCNN (cls-agn) 			& 34.3 			& 36.2 			& 39.3 			& 23.1 & 25.0 & 29.5 & 1.8 \\ 
    ORCNN 						& 35.1 			& 37.6 			& 39.4 			& 23.7 & 25.8 & 29.9 & \textbf{2.0} \\ 
    \multicolumn{8}{c}{} \hfill
  \end{tabular}
  \egroup
\end{table}

\subsection{\emph{D2S amodal}}

\emph{D2S} is a recent dataset of supermarket products that should capture the setting and needs of industrial applications. In particular, the low complexity of the training set with respect to its size and image attributes (homogeneous background, no clutter) makes it necessary to use data augmentation. Since only minor occlusions are present in the training set, the generation of reasonable artificial images is straight-forward. In this work, we provide additional amodal annotations for all images of \emph{D2S} to obtain \emph{D2S amodal}. The annotations contain the category, amodal mask, and additional visible and invisible masks for occluded objects. For images where the amodal masks are reaching out of the image boundary, a zero-padding is used such that all amodal masks are fully contained in the image.

\begin{figure*}
  \centering
  \includegraphics[width=0.24\textwidth]{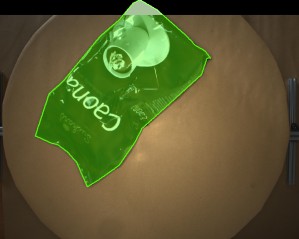} 
  \includegraphics[width=0.24\textwidth]{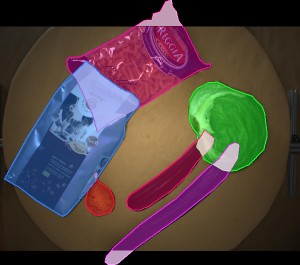} 
  \includegraphics[width=0.24\textwidth]{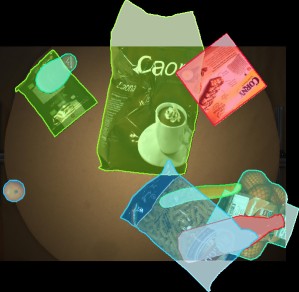} 
  \includegraphics[width=0.24\textwidth]{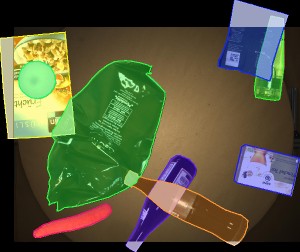} 
  \caption{\textbf{D2S amodal ground truth.}
    Ground truth annotations for different splits of the \emph{D2S amodal} dataset.
    (\emph{From left to right}) train, val, aug, modal aug}
  \label{fig:d2s_gt_annots}
\end{figure*}

\paragraph{Splits.} As the \emph{D2S amodal} training split only contains minor occlusions, it is not suitable to train semantic amodal models that predict occlusion masks also for moderately to heavily occluded objects. Hence, we use data augmentation similar to the data augmentation in \cite{d2s2018}. \emph{D2S amodal augmented} consists of 1562 augmented images, where only objects from \emph{D2S train} that do not reach out of the image boundary have been used for augmentation. The \emph{D2S amodal train} set is then the combination of \emph{D2S amodal training\_rot0} with \emph{D2S amodal augmented} splits resulting in a total of 2000 images.

To evaluate if an amodal model can be trained only from modal annotated data, we augmented another 2000 images in the same way, but using the modal annotations from \emph{D2S training} to obtain \emph{D2S amodal modal augmented}. For all splits of \emph{D2S amodal}, the statistics are given in \tabref{tbl:d2s_stats}. As is common for \emph{D2S}, the number of images containing occlusion are a lot higher for the validation and test set compared to the train\_rot0 set. Using the augmented set, one can create an overall train split that contains many objects and even a higher frequency and average per-object occlusion rate than in the validation and test splits. Exemplary ground truth annotations for different splits are shown in \figref{fig:d2s_gt_annots}.

\begin{table}[ht]
  \caption{\textbf{\emph{D2S amodal} splits.} Image and occlusion statistics for the splits used with \emph{D2S amodal} from top to bottom: number of images, number of images with at least one occluded object, rate of images that contain any occlusion, total number of objects, number of occluded objects, rate of objects that are occluded, average occlusion rate per object region for all objects, average occlusion rate per object only for occluded objects.}
  \label{tbl:d2s_stats}
  \centering
  \small
  \bgroup
  \def\arraystretch{1.2} 
  \setlength{\tabcolsep}{4pt}
  \begin{tabular}{l | r r r r r r} \hline
    split name               & aug & train\_rot0 & train & val & test & modal aug  \\ \hline \hline
    num imgs                 & 1562 & 438 & 2000 & 3600 & 13020 & 2000 \\
    num imgs w/ occl         & 1507	& 57 & 1564	& 2520 & 8976 & 1930 \\
    img OR [\%]              & 96 & 13 & 78 & 70 &	68 & 96 \\
    num objs                 & 12376 & 690 & 13066 & 15654 & 49914 & 15851 \\
    num objs occl            & 8798 & 66 & 8864 & 7473 & 21966 & 11302 \\
    obj OR [\%]              & 71 & 9 & 68 &	47 & 44	& 71 \\
    avg OR / reg (all) [\%]  & 23 & 0 & 22 & 8 & 8 & 23 \\
    avg OR / reg (occl) [\%] & 33 & 4 & 33 & 18 & 18 & 33 \\ \hline
    \multicolumn{7}{c}{} \hfill
  \end{tabular}
  \egroup
\end{table}
\paragraph{Architectures.} As before, AmodalMRCNN is the architecture of \cite{he_2017_mask} trained to predict amodal masks. For ORCNN, in order to test the influence of adding the visible and invisible mask losses, we propose four different architectures: ORCNN is using a loss for each of the three mask types (amodal, visible and invisible). For ORCNN (w/o $L_{IV}$) and ORCNN (w/o $L_V$) the loss for invisible or visible mask prediction is switched off, respectively. ORCNN (independent) is a model including $L_{IV}$ and $L_{V}$ but where the two losses are not propagated to the amodal mask head nor to the RoI feature extraction part.

\begin{figure*}
  \centering
  \includegraphics[width=0.19\textwidth]{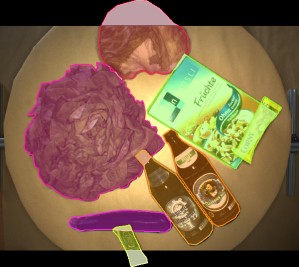}
  \includegraphics[width=0.19\textwidth]{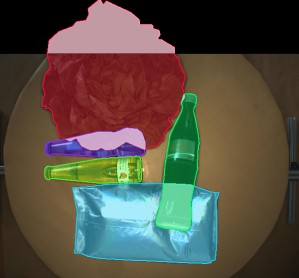}
  \includegraphics[width=0.19\textwidth]{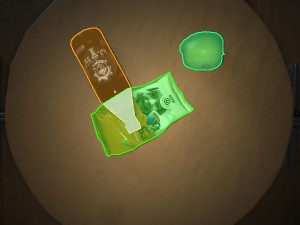}
  \includegraphics[width=0.19\textwidth]{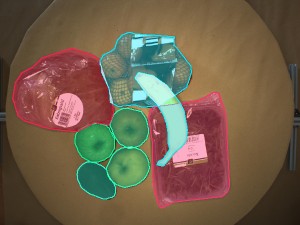}
  \includegraphics[width=0.19\textwidth]{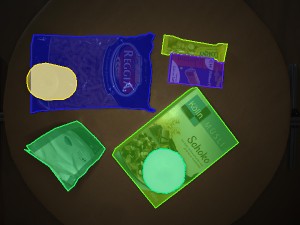}
  \includegraphics[width=0.19\textwidth]{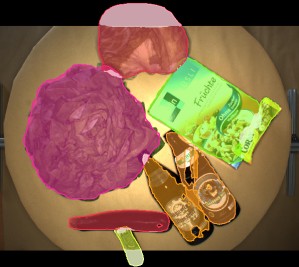}
  \includegraphics[width=0.19\textwidth]{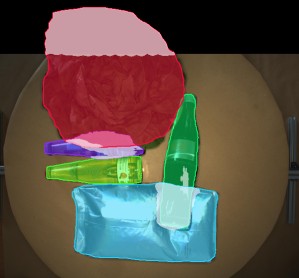}
  \includegraphics[width=0.19\textwidth]{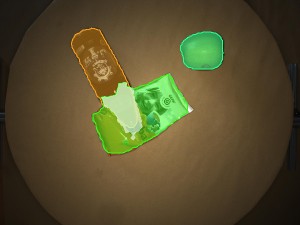}
  \includegraphics[width=0.19\textwidth]{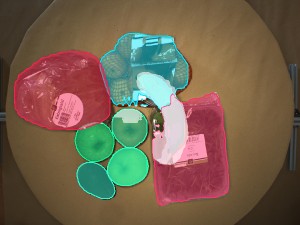}
  \includegraphics[width=0.19\textwidth]{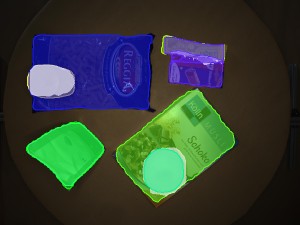}
  \caption{\textbf{\emph{D2S amodal} results.} (\emph{Top}) Ground truth annotations for images of the validation set; (\emph{bottom}) exemplary qualitative results of our ORCNN model trained on the \texttt{amodal\_train} split. Predicted invisible masks are indicated by a white overlap. In most cases the results are promising, especially for objects lying on top of each other or reaching beyond the original image boundary}
  \label{fig:d2s_results}
\end{figure*}

The qualitative results of ORCNN in \figref{fig:d2s_results} are very promising. The model predicts occlusions correctly in many cases, especially for objects lying completely on top of another one or objects reaching out of the image boundary. This shows that the model is able to learn the common shape of object classes.

Nevertheless, the quantitative analysis given in \tabref{tbl:d2s_results} shows that the prediction of invisible masks is very difficult. On the one hand, if the ground truth invisible masks are small, then small differences of the invisible mask proposal already lead to a small IoU value. On the other hand, if the ground truth invisible mask is large, it is very difficult for the model to generate the correct shape of the invisible mask prediciton, again leading to a small IoU value. In both cases if the IoU value is below 0.5 this leads to a false positive prediction and possibly a false negative as the corresponding ground truth is not matched.

ORCNN has the ability to predict invisible masks as the difference of predicted amodal and visible masks in a single forward pass. This comes at the cost that amodal masks are slightly worse than for AmodalRCNN.

\begin{figure*}
  \centering
  \includegraphics[width=0.24\textwidth]{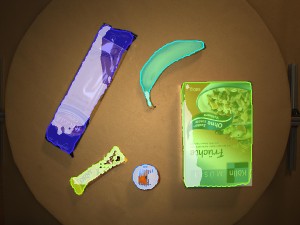}
  \includegraphics[width=0.24\textwidth]{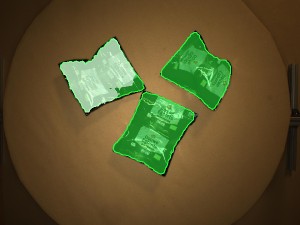}
  \includegraphics[width=0.24\textwidth]{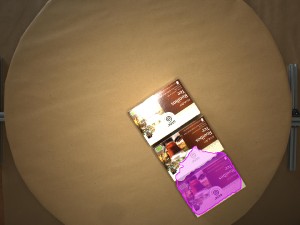}
  \includegraphics[width=0.24\textwidth]{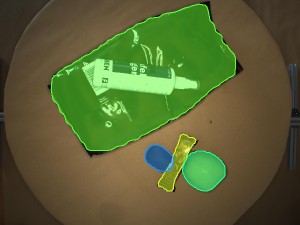}
  \caption{\textbf{D2S failure cases.}
    Exemplary qualitative results where our ORCNN model fails. Predicted invisible masks are indicated by a white overlap. (\emph{From left to right}) (\emph{1 and 2}) occlusion is falsely predicted possibly due to reflections or lighting changes; (\emph{3}) the amodal and invisible mask of this instance is extended into the neighboring object of the same class (only this instance is visualized); (\emph{4}) the occlusion caused by the plastic bottle is not detected.}
  \label{fig:d2s_failure_cases}
\end{figure*}

Some failure cases of ORCNN on \emph{D2S amodal} are shown in \figref{fig:d2s_failure_cases}. False positive occlusion predictions are often caused by reflections or lighting changes. When objects are lying next to each other and touching, the amodal and invisible masks are sometimes extended into the neighboring instance. For other cases, occlusions are not predicted at all.

In summary, ORCNN is the best compromise for the prediction of visible and invisible masks at the same time (highest  $AP_{\mathit{AV}}$ and  $AP_{\mathit{V}}$). This comes at the cost of a slightly lower $AP_{\mathit{A}}$ value compared to a model that only predicts amodal masks, like AmodalMRCNN. 
Interestingly, for \emph{D2S amodal} ORCNN (cls-agn) outperforms the ORCNN model that predicts a class specific mask. This result contrasts with what we obtained for \emph{COCOA} and \emph{COCOA cls}.

\figref{fig:d2s_results} also shows the result of ORCNN when training only on artificially augmented data obtained from \emph{D2S train}, \emph{D2S amodal modal augmented}. The model performs only slightly worse than ORCNN trained on \emph{D2S amodal train}. This is in difference to \emph{COCOA}, where data augmentation using modal annotations did not help and led to significantly worse results than training on amodal data. 
\begin{table}[ht]
  \caption{\textbf{\emph{D2S amodal} results.} Models marked with (cls-agn) are using class-agnostic mask prediction heads. ORCNN (modal aug) is a ORCNN only trained on \emph{D2s amodal modal augmented}. All values are calculated for the \emph{D2S amodal validation} set.}
  \label{tbl:d2s_results}
  \centering
  \small
  \bgroup
  \def\arraystretch{1.2} 
  \setlength{\tabcolsep}{4pt}
  \begin{tabular}{l | r r r | r r r r} \hline
    & \multicolumn{3}{|c|}{all} & \multicolumn{4}{c}{occluded} \\
    dataset \& model & $AP_{\mathit{AV}}$ & $AP_{\mathit{A}}$ & $AP_{\mathit{V}}$ & $AP_{\mathit{AV}}$ & $AP_{\mathit{A}}$ & $AP_{\mathit{V}}$ & $AP^{0.5}_{\mathit{IV}}$ \\ \hline \hline
    AmodalMRCNN-101 (cls-agn) & 63.4 & 72.2 & 64.8 & 48.0 & 62.2 & 50.8 & - \\
    AmodalMRCNN-101           & \textbf{63.8} & 72.6 & 65.3 & 48.7 & \textbf{63.0} & 51.6 & - \\
    ORCNN (w/o $L_{IV}$)      & 34.0 & 68.6 & 34.7 & 26.8 & 58.9 & 28.0 & 0.1  \\
    ORCNN (w/o $L_{V}$)       & 11.0 & 66.1 & 11.1 & 7.5 & 56.7 & 7.6 & 5.8 \\ 
    ORCNN (independent)       & 39.5 & \textbf{72.8} & 40.3 & 31.9 & 62.9 & 33.2 & 0.1 \\
    ORCNN (cls-agn)           & 62.3 & 65.2 & \textbf{71.0} & \textbf{52.5} & 55.2 & \textbf{65.3} & \textbf{14.7}  \\ 
    ORCNN                     & 58.9 & 62.1 & 66.2 & 48.1 & 51.2 & 58.9 & 8.7 \\ \hline 
    ORCNN (modal aug)         & 56.3 & 59.9 & 62.6 & 47.9 & 51.0 & 57.6 & 7.8 \\ \hline 
    \multicolumn{8}{c}{} \hfill
  \end{tabular}
  \egroup
\end{table}

\section{Conclusion}

We proposed an end-to-end trainable, instance-aware model for semantic amodal segmentation. Our model, ORCNN, simultaneously predicts amodal, visible and invisible masks, and the category of each instance in a single forward pass.
By merging annotations of \emph{COCO} with \emph{COCOA}, we obtain a category-specific semantic amodal dataset based on \emph{COCO} images \emph{COCOA cls}.
We provide semantic amodal ground truth for \emph{D2S} splits resulting in \emph{D2S amodal}.
ORCNN outperforms previous work on \emph{COCOA} and sets strong benchmarks on \emph{COCOA cls} as well as \emph{D2S amodal}.
Our experiments and results show that it is possible to predict the invisible masks of occluded objects even in areas without any visual cue. Thus, our model can indeed \emph{learn to see the invisible}.

The results show that the prediction of amodal and in particular invisible masks is a difficult task that needs further research to reduce the number of false positive predictions of occlusions.


\appendix
\section*{Appendix}

The following will be provided as supplementary:
\begin{itemize}
  \item Description of training configuration.
  \item Qualitative Results \emph{COCOA}.
  \item Qualitative Results \emph{D2S amodal}.
\end{itemize}

\section{Training Setup}

As stated in the paper, for the training of AmodalMRCNN as well as ORCNN we used the Detectron \cite{Detectron2018} framework built on caffe2\footnote{https://github.com/caffe2/caffe2} / caffe \cite{jia2014caffe}.

\paragraph{Batch- and image-size.} For all trainings 2 NVIDIA GTX 1080 Ti GPUs were used. On \emph{D2S amodal} we used a batch-size of one image per GPU, while on \emph{COCOA} and \emph{COCOA cls} two images per GPU were used in a batch. The batch size of regions of interest (RoIs) per image was set to 256. The maximum number of RoIs per image before the non-maximum suppression in the region proposal network was set to 1000 per feature pyramid level. The images were scaled such that the shorter side of each image was set to a maximum of 800 pixels while the longer side was not exceeding 1111 pixels.

\paragraph{Finetuning.} Each training was using the weights from a \emph{COCO}-pretrained model as provided by Detectron. For the case of \emph{COCOA} and \emph{D2S amodal} the final output layers that are class-specific had to be initialized randomly as the number of classes and their semantic meaning did not fit to the number of classes of \emph{COCO}.

\paragraph{Solver settings.} Training was performed for 10000 iterations with a base learning rate of 0.0025 and a weight decay of 0.0001. The same warm-up period and settings as described in the Mask R-CNN paper \cite{he_2017_mask} were used for the learning rate. The learning rate was multiplied by $\gamma = 0.1$ at 6000 and 8000 iterations, respectively.

\section{Qualitative Results}

In this section qualitative results are shown on \emph{COCOA no stuff}, \emph{COCOA cls} and \emph{D2S amodal}. ORCNN, a model that can predict amodal, invisible and visible masks simultaneously, is compared against AmodalMRCNN that is only capable to predict amodal masks.

\subsection{\emph{COCOA}}
\begin{figure}
  \begin{center}
    \begin{tabular}{ c c c c }
      original image & ground truth & AmodalMRCNN & ORCNN \\
      \includegraphics[width=0.225\textwidth]{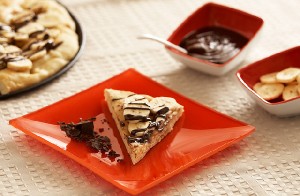} &
      \includegraphics[width=0.225\textwidth]{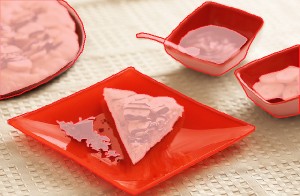} & 
      \includegraphics[width=0.225\textwidth]{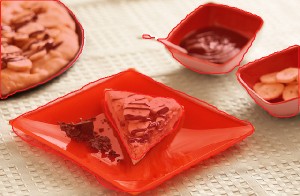} &
      \includegraphics[width=0.225\textwidth]{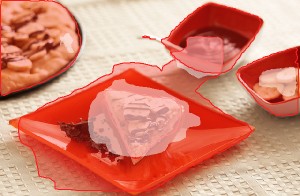} \\
      \includegraphics[width=0.225\textwidth]{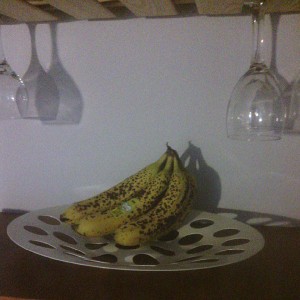} &
      \includegraphics[width=0.225\textwidth]{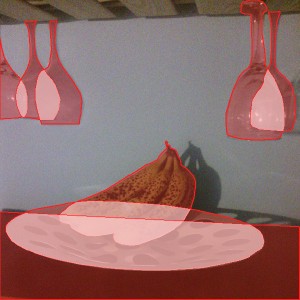} & 
      \includegraphics[width=0.225\textwidth]{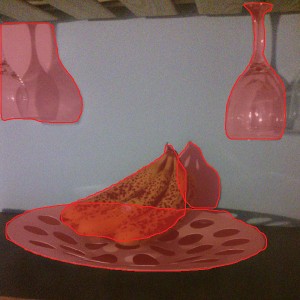} &
      \includegraphics[width=0.225\textwidth]{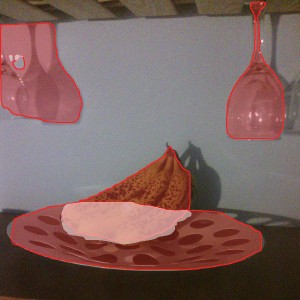} \\
      \includegraphics[width=0.225\textwidth]{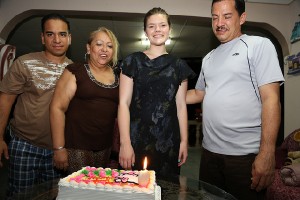} &
      \includegraphics[width=0.225\textwidth]{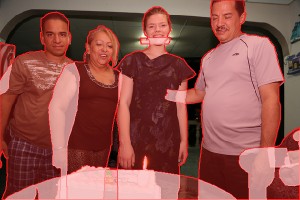} & 
      \includegraphics[width=0.225\textwidth]{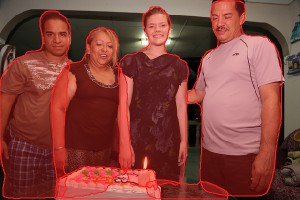} &
      \includegraphics[width=0.225\textwidth]{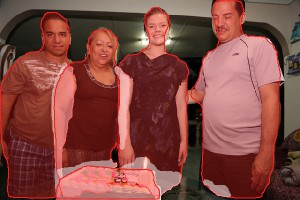} \\
      \includegraphics[width=0.225\textwidth]{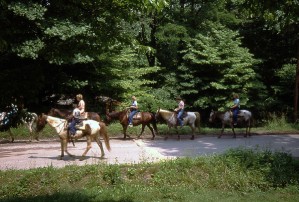} &
      \includegraphics[width=0.225\textwidth]{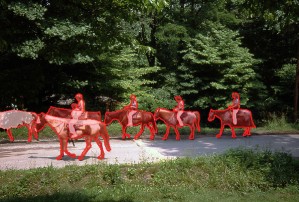} & 
      \includegraphics[width=0.225\textwidth]{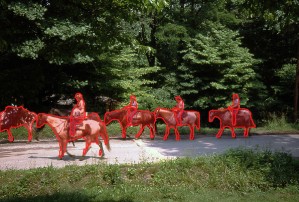} &
      \includegraphics[width=0.225\textwidth]{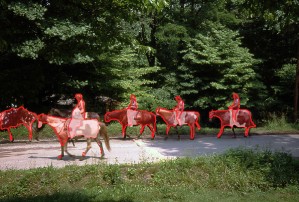} \\
      \includegraphics[width=0.225\textwidth]{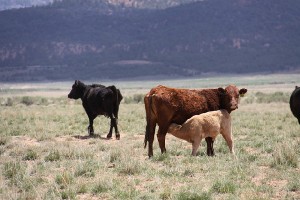} &
      \includegraphics[width=0.225\textwidth]{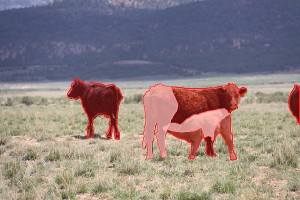} & 
      \includegraphics[width=0.225\textwidth]{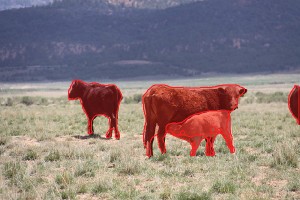} &
      \includegraphics[width=0.225\textwidth]{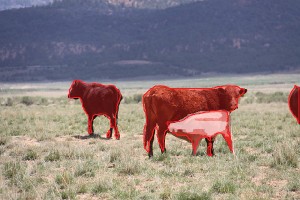} \\
      \includegraphics[width=0.225\textwidth]{COCO_val2014_000000473541_orcnn.jpg} &
      \includegraphics[width=0.225\textwidth]{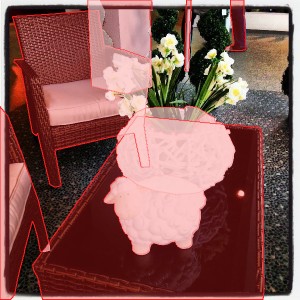} & 
      \includegraphics[width=0.225\textwidth]{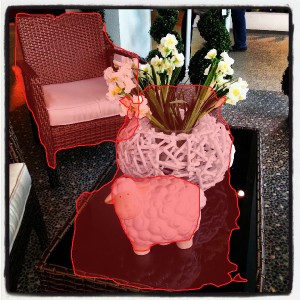} &
      \includegraphics[width=0.225\textwidth]{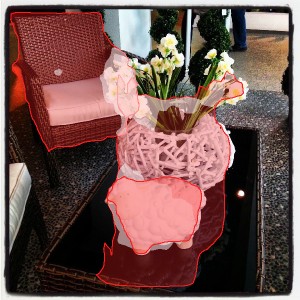} \\
    \end{tabular}
    \caption{\textbf{COCOA no stuff results.}
      \textit{From left to right:} input image, ground truth annotations, results of AmodalMRCNN, results of ORCNN. Note that for all images the minimal score of results was set to 0.8. The ordering of the result overlays might be different to the ordering of the ground truth annotation overlays. This will be changed for the camera ready version of the paper.}
    \label{fig:resCOCOAnostuff}
  \end{center}
\end{figure}

\begin{figure}
  \begin{center}
    \begin{tabular}{ c c c c }
      original image & ground truth & AmodalMRCNN & ORCNN \\
      \includegraphics[width=0.225\textwidth]{COCO_val2014_000000223335_orcnn.jpg} &
      \includegraphics[width=0.225\textwidth]{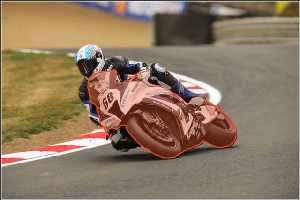} & 
      \includegraphics[width=0.225\textwidth]{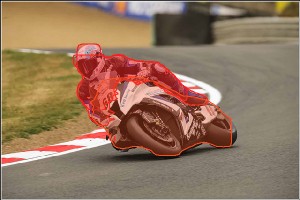} &
      \includegraphics[width=0.225\textwidth]{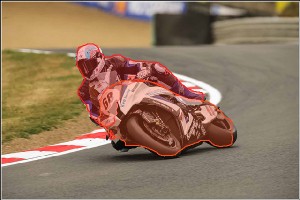} \\
      \includegraphics[width=0.225\textwidth]{COCO_val2014_000000277197_orcnn.jpg} &
      \includegraphics[width=0.225\textwidth]{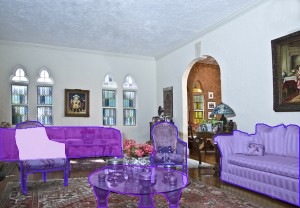} & 
      \includegraphics[width=0.225\textwidth]{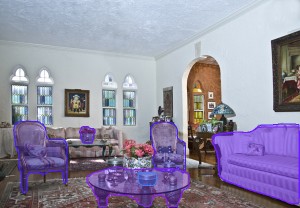} &
      \includegraphics[width=0.225\textwidth]{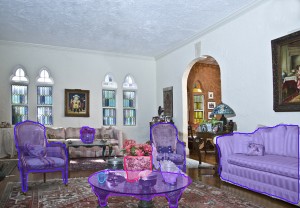} \\
      \includegraphics[width=0.225\textwidth]{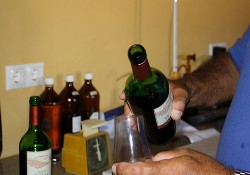} &
      \includegraphics[width=0.225\textwidth]{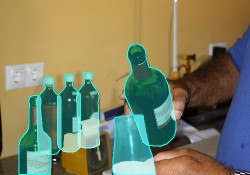} & 
      \includegraphics[width=0.225\textwidth]{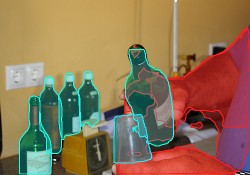} &
      \includegraphics[width=0.225\textwidth]{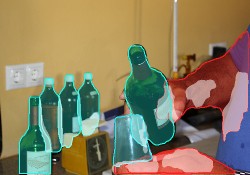} \\
      \includegraphics[width=0.225\textwidth]{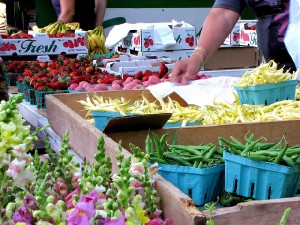} &
      \includegraphics[width=0.225\textwidth]{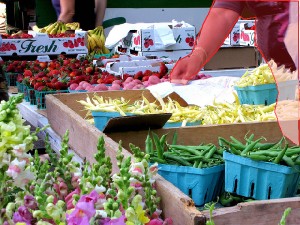} & 
      \includegraphics[width=0.225\textwidth]{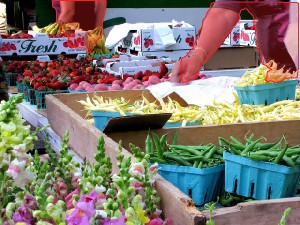} &
      \includegraphics[width=0.225\textwidth]{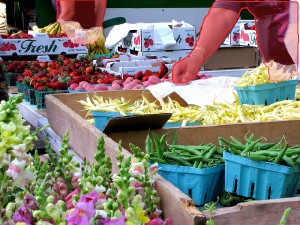} \\
      \includegraphics[width=0.225\textwidth]{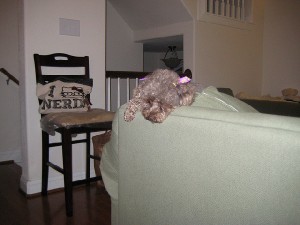} &
      \includegraphics[width=0.225\textwidth]{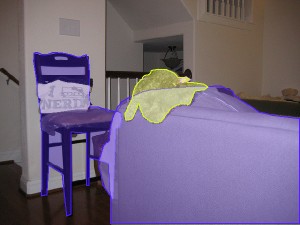} & 
      \includegraphics[width=0.225\textwidth]{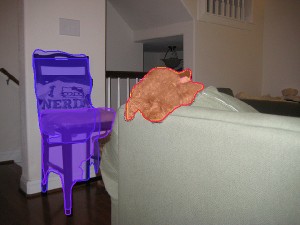} &
      \includegraphics[width=0.225\textwidth]{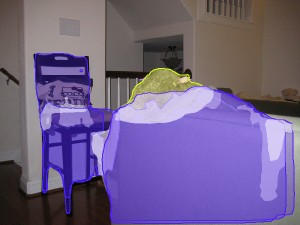} \\
      \includegraphics[width=0.225\textwidth]{COCO_val2014_000000579697_orcnn.jpg} &
      \includegraphics[width=0.225\textwidth]{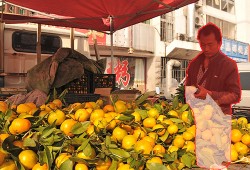} & 
      \includegraphics[width=0.225\textwidth]{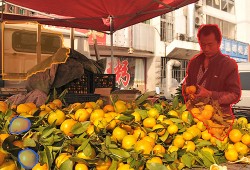} &
      \includegraphics[width=0.225\textwidth]{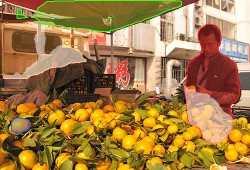} \\
      \includegraphics[width=0.225\textwidth]{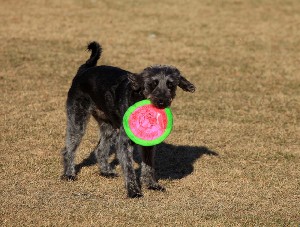} &
      \includegraphics[width=0.225\textwidth]{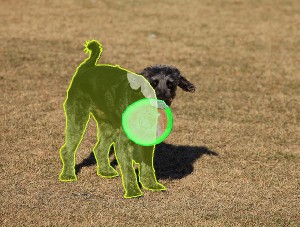} & 
      \includegraphics[width=0.225\textwidth]{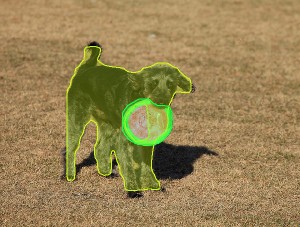} &
      \includegraphics[width=0.225\textwidth]{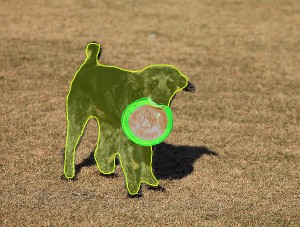} \\
    \end{tabular}
    \caption{\textbf{COCOA cls results.}
      \textit{From left to right:} input image, ground truth annotations, results of AmodalMRCNN, results of ORCNN. Note that for all images the minimal score of results was set to 0.5. The ordering of the result overlays might be different to the ordering of the ground truth annotation overlays. This will be changed for the camera ready version of the paper.}
    \label{fig:resCOCOAcls}
  \end{center}
\end{figure}

Qualitative results for \emph{COCOA no stuff} and \emph{COCOA cls} are shown in \figref{fig:resCOCOAnostuff} and \figref{fig:resCOCOAcls}, respectively. Please note that the ground truth annotations are sometimes wrong or incomplete. Hence, the models sometimes find annotations correctly that have not been annotated in the ground truth. This is a common problem of large-scale datasets that have are annotated using Amazon Mechanical Turck such as \emph{COCO}. For datasets of this size, variety and complexity it is not feasible to control all ground truth annotations. In the case of \emph{COCOA cls}, missing annotations are also generated by our merging strategy. We do not show qualitative results for \emph{COCOA} as the ground truth and results for \emph{stuff} regions make it very difficult to visualize and interpret the results.

Generally, we observe that the predicted amodal regions are a little bit better for AmodalMRCNN compared to ORCNN. This is also confirmed by the quantitative evaluation in the paper. However, the occlusion predictions are often very promising and better then the relatively low $AP_{\mathrm{IV}}$ values indicate.

\subsection{\emph{D2S}}

\begin{figure}
  \begin{center}
    \begin{tabular}{ c c c c }
      original image & ground truth & AmodalMRCNN & ORCNN \\
      \includegraphics[width=0.22\textwidth]{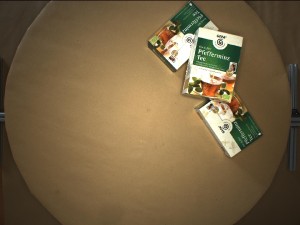} &
      \includegraphics[width=0.22\textwidth]{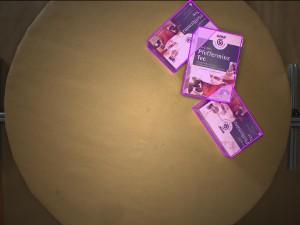} & 
      \includegraphics[width=0.22\textwidth]{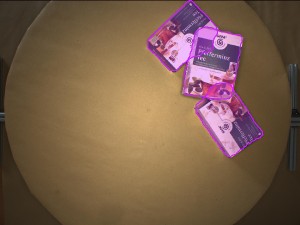} &
      \includegraphics[width=0.22\textwidth]{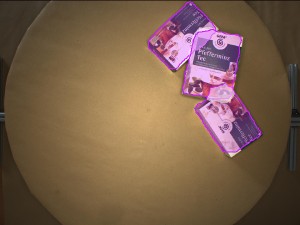} \\
      \includegraphics[width=0.22\textwidth]{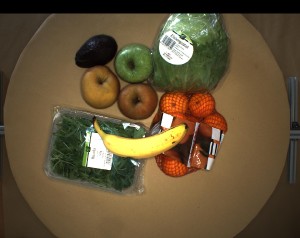} &
      \includegraphics[width=0.22\textwidth]{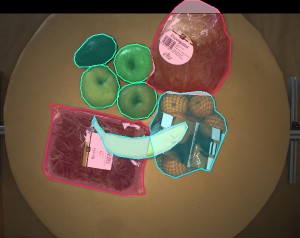} & 
      \includegraphics[width=0.22\textwidth]{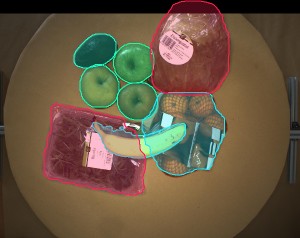} &
      \includegraphics[width=0.22\textwidth]{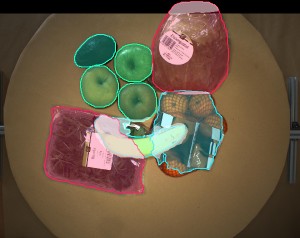} \\
      \includegraphics[width=0.22\textwidth]{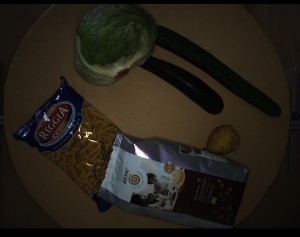} &
      \includegraphics[width=0.22\textwidth]{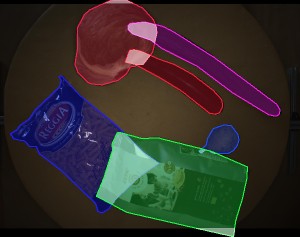} & 
      \includegraphics[width=0.22\textwidth]{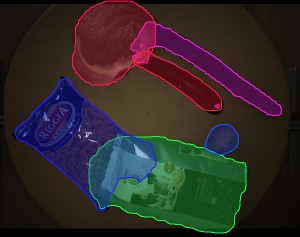} &
      \includegraphics[width=0.22\textwidth]{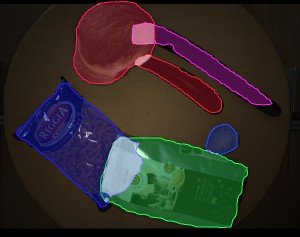} \\
      \includegraphics[width=0.22\textwidth]{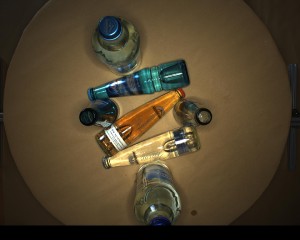} &
      \includegraphics[width=0.22\textwidth]{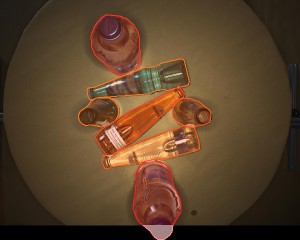} & 
      \includegraphics[width=0.22\textwidth]{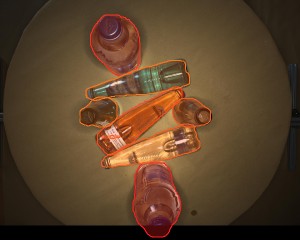} &
      \includegraphics[width=0.22\textwidth]{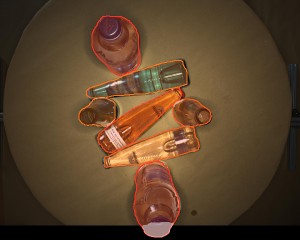} \\
      \includegraphics[width=0.22\textwidth]{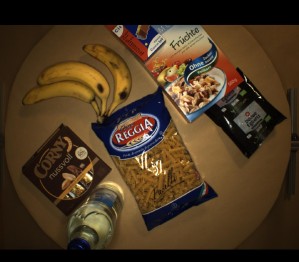} &
      \includegraphics[width=0.22\textwidth]{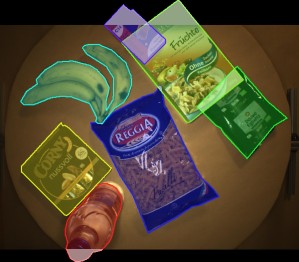} & 
      \includegraphics[width=0.22\textwidth]{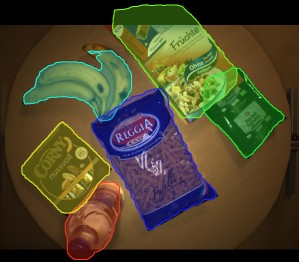} &
      \includegraphics[width=0.22\textwidth]{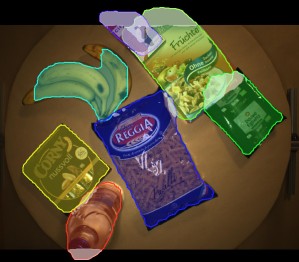} \\
      \includegraphics[width=0.22\textwidth]{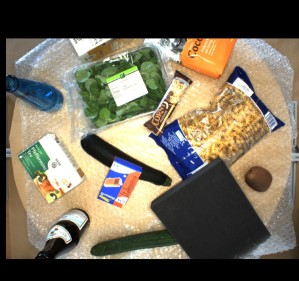} &
      \includegraphics[width=0.22\textwidth]{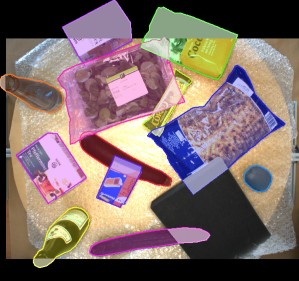} & 
      \includegraphics[width=0.22\textwidth]{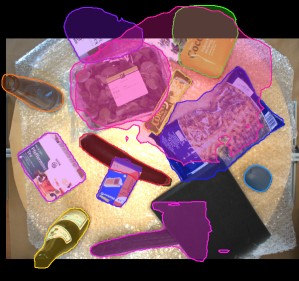} &
      \includegraphics[width=0.22\textwidth]{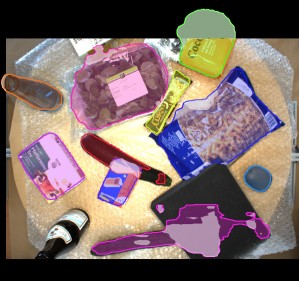} \\
      \includegraphics[width=0.22\textwidth]{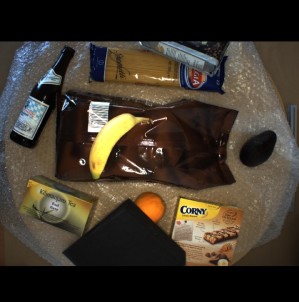} &
      \includegraphics[width=0.22\textwidth]{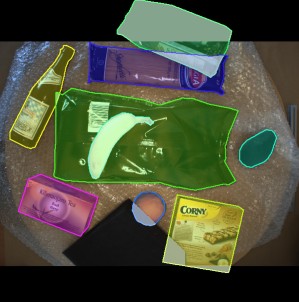} & 
      \includegraphics[width=0.22\textwidth]{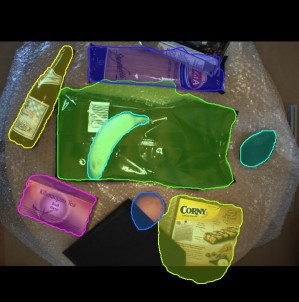} &
      \includegraphics[width=0.22\textwidth]{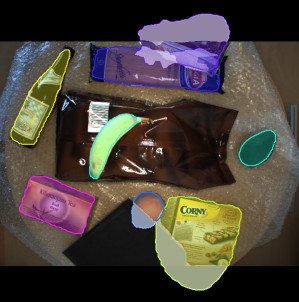} \\
    \end{tabular}
    \caption{\textbf{D2S amodal results.}
      \textit{From left to right:} input image, ground truth annotations, results of AmodalMRCNN, results of ORCNN. Note that for all images the minimal score of results was set to 0.8. The ordering of the result overlays might be different to the ordering of the ground truth annotation overlays.}
    \label{fig:resD2Samodal}
  \end{center}
\end{figure}

For \emph{D2S amodal} we show qualitative examples for the class-agnostic mask versions of AmodalMRCNN and ORCNN in \figref{fig:resD2Samodal}. It is difficult to see a qualitative difference between the amodal masks of AmodalMRCNN and ORCNN. The predicted invisible masks of ORCNN are often at the correct location but their IoU with the ground truth invisible masks is rather low. This explains the low $AP_{\mathrm{IV}}$ values.

\clearpage

\bibliographystyle{splncs03}
\bibliography{egbib}

\begin{thebibliography}{10}
\providecommand{\url}[1]{\texttt{#1}}
\providecommand{\urlprefix}{URL }

\bibitem{chen2015multi}
Chen, Y.T., Liu, X., Yang, M.H.: Multi-instance object segmentation with
  occlusion handling. In: Conference on Computer Vision and Pattern
  Recognition. pp. 3470--3478 (2015)

\bibitem{everingham_2015_pascal}
Everingham, M., Eslami, S.M.A., Gool, L.J.V., Williams, C.K.I., Winn, J.M.,
  Zisserman, A.: The pascal visual object classes challenge: {A} retrospective.
  International Journal of Computer Vision  111(1),  98--136 (2015),
  \url{https://doi.org/10.1007/s11263-014-0733-5}

\bibitem{d2s2018}
Follmann, P., B\"ottger, T., H\"artinger, P., K\"onig, R., Ulrich, M.: {MVT}ec
  {D2S}: {D}ensely {S}egmented {S}upermarket {D}ataset. CoRR  abs/1804.08292
  (2018), \url{https://arxiv.org/abs/1804.08292}

\bibitem{Detectron2018}
Girshick, R., Radosavovic, I., Gkioxari, G., Doll\'{a}r, P., He, K.: Detectron.
  \url{https://github.com/facebookresearch/detectron} (2018)

\bibitem{guo2012beyond}
Guo, R., Hoiem, D.: Beyond the line of sight: labeling the underlying surfaces.
  In: European Conference on Computer Vision. pp. 761--774. Springer (2012)

\bibitem{gupta2013perceptual}
Gupta, S., Arbelaez, P., Malik, J.: Perceptual organization and recognition of
  indoor scenes from rgb-d images. In: Conference on Computer Vision and
  Pattern Recognition. pp. 564--571. IEEE (2013)

\bibitem{he_2017_mask}
He, K., Gkioxari, G., Dollar, P., Girshick, R.: Mask {R-CNN}. In: International
  Conference on Computer Vision. pp. 1059--1067 (2017)

\bibitem{he2016deep}
He, K., Zhang, X., Ren, S., Sun, J.: Deep residual learning for image
  recognition. In: Conference on Computer Vision and Pattern Recognition. pp.
  770--778 (2016)

\bibitem{jia2014caffe}
Jia, Y., Shelhamer, E., Donahue, J., Karayev, S., Long, J., Girshick, R.,
  Guadarrama, S., Darrell, T.: Caffe: Convolutional architecture for fast
  feature embedding. arXiv preprint arXiv:1408.5093  (2014)

\bibitem{kar2015amodal}
Kar, A., Tulsiani, S., Carreira, J., Malik, J.: Amodal completion and size
  constancy in natural scenes. In: International Conference on Computer Vision.
  pp. 127--135 (2015)

\bibitem{kellman2013perceptual}
Kellman, P.J., Massey, C.M.: Perceptual learning, cognition, and expertise. In:
  Psychology of learning and motivation, vol.~58, pp. 117--165. Elsevier (2013)

\bibitem{lehar1999gestalt}
Lehar, S.: Gestalt isomorphism and the quantification of spatial perception.
  Gestalt theory  21,  122--139 (1999)

\bibitem{li2016iterative}
Li, K., Hariharan, B., Malik, J.: Iterative instance segmentation. In:
  Conference on Computer Vision and Pattern Recognition. pp. 3659--3667 (2016)

\bibitem{li2016amodal}
Li, K., Malik, J.: Amodal instance segmentation. In: European Conference on
  Computer Vision Submission. pp. 677--693. Springer (2016)

\bibitem{li_2016_fully}
Li, Y., Qi, H., Da, J., Ji, X., Wei, Y.: Fully convolutional instance-aware
  semantic segmentation. In: Conference on Computer Vision and Pattern
  Recognition. pp. 2359--2367 (2017)

\bibitem{lin_2014_coco}
Lin, T., Maire, M., Belongie, S.J., Hays, J., Perona, P., Ramanan, D.,
  Doll{\'{a}}r, P., Zitnick, C.L.: Microsoft {COCO:} common objects in context.
  In: European Conference on Computer Vision Submission. pp. 740--755 (2014),
  \url{https://doi.org/10.1007/978-3-319-10602-1_48}

\bibitem{pinhero2015}
Pinheiro, P.O., Collobert, R., Dollar, P.: Learning to segment object
  candidates. In: Advances in Neural Information Processing Systems 28, pp.
  1990--1998 (2015),
  \url{http://papers.nips.cc/paper/5852-learning-to-segment-object-candidates.pdf}

\bibitem{silberman2014contour}
Silberman, N., Shapira, L., Gal, R., Kohli, P.: A contour completion model for
  augmenting surface reconstructions. In: European Conference on Computer
  Vision. pp. 488--503. Springer (2014)

\bibitem{yang2012layered}
Yang, Y., Hallman, S., Ramanan, D., Fowlkes, C.C.: Layered object models for
  image segmentation. IEEE Transactions on Pattern Analysis and Machine
  Intelligence  34(9),  1731--1743 (2012)

\bibitem{zhu2017semantic}
Zhu, Y., Tian, Y., Metaxas, D., Dollar, P.: Semantic amodal segmentation. In:
  Conference on Computer Vision and Pattern Recognition. pp. 1464--1472 (2017)

\end{thebibliography}
\end{document}